%% file: root.tex
\documentclass[letterpaper, 10 pt, conference]{ieeeconf}  

\IEEEoverridecommandlockouts                              

\usepackage{graphics} 
\usepackage{subcaption}
\usepackage{epsfig} 
\usepackage{mathptmx} 
\usepackage{amsmath} 
\usepackage{amssymb}  
\usepackage[T1]{fontenc}
\usepackage[colorlinks, citecolor=red, linkcolor=black]{hyperref}
\usepackage{cite}
\usepackage{makecell}
\usepackage{multirow}

\title{\LARGE \bf
FastOcc: Accelerating 3D Occupancy Prediction by Fusing the 2D Bird's-Eye View and Perspective View
}

\author{{\normalsize Jiawei Hou$^{1*}$, Xiaoyan Li$^{2*}$, Wenhao Guan$^{1*}$, Gang Zhang$^{3}$, Di Feng$^{3}$, Yuheng Du$^{1}$ Xiangyang Xue$^{1}$, and Jian Pu$^{4\dag}$}
\thanks{$^{*}$ These authors contributed equally to this work}
\thanks{$^{\dag}$ Corresponding author}
\thanks{$^{1}$  School of Computer Science, Fudan University, Shanghai, China
        {\tt\small \{jwhou23, whguan21, yhdu22\}@m.fudan.edu.cn}
        , xyxue@fudan.edu.cn}%
\thanks{$^{2}$  
        Faculty of Information Technology, Beijing University of Technology, Beijing 100124, China
        {\tt\small xiaoyan.li@bjut.edu.cn}
        }%
\thanks{$^{3}$  Mogo Auto Intelligence and Telematics Information Technology Co., Ltd
        {\tt\small zhanggang11021136@gmail.com}
        }%
\thanks{$^{4}$  Institute of Science and Technology for Brain-Inspired Intelligence, Fudan University, Shanghai, China
        {\tt\small jianpu@fudan.edu.cn}
        }%
}

\begin{document}

\maketitle
\thispagestyle{empty}
\pagestyle{empty}

\begin{abstract}


In autonomous driving, 3D occupancy prediction outputs voxel-wise status and semantic labels for more comprehensive understandings of 3D scenes compared with traditional perception tasks, such as 3D object detection and bird's-eye view (BEV) semantic segmentation. Recent researchers have extensively explored various aspects of this task, including view transformation techniques, ground-truth label generation, and elaborate network design, aiming to achieve superior performance. However, the inference speed, crucial for running on an autonomous vehicle, is neglected. To this end, a new method, dubbed FastOcc, is proposed. By carefully analyzing the network effect and latency from four parts, including the input image resolution, image backbone, view transformation, and occupancy prediction head, it is found that the occupancy prediction head holds considerable potential for accelerating the model while keeping its accuracy. Targeted at improving this component, the time-consuming 3D convolution network is replaced with a novel residual-like architecture, where features are mainly digested by a lightweight 2D BEV convolution network and compensated by integrating the 3D voxel features interpolated from the original image features. Experiments on the Occ3D-nuScenes benchmark demonstrate that our FastOcc achieves state-of-the-art results with a fast inference speed.

\end{abstract}


\input{sections/introduction}

\input{sections/related_work}

\input{sections/occupancy_prediction}

\input{sections/experiment}

\input{sections/conclusions}

\section*{ACKNOWLEDGMENT}
This paper is supported in part by Shanghai Platform for Neuromorphic and AI Chip under Grant 17DZ2260900 (NeuHelium).

\addtolength{\textheight}{-0cm}   






\bibliographystyle{IEEEtran}
\bibliography{IEEEabrv, references}

\end{document}

%% file: sections/introduction.tex
\section{INTRODUCTION}

Understanding the 3D geometry and semantic information of the surrounding scene is a crucial problem for autonomous driving. Recently, camera-based perception methods have gained widespread concerns due to their lower costs than the LiDAR-based methods. Several approaches have reached remarkable achievements in the 3D perception tasks, such as 3D object detection~\cite{huangBEVDetHighperformanceMulticamera2021, yangBEVFormerV2Adapting2022, wangExploringObjectCentricTemporal2023}, bird's-eye-view (BEV) semantic segmentation~\cite{pengBEVSegFormerBirdEye2022, philionLiftSplatShoot2020, zhouCrossviewTransformersRealtime2022, zhuNeMONeuralMap2023}, \emph{etc}. However, tasks such as 3D object detection are plagued by the long-tail issue and have difficulty recognizing objects with arbitrary shapes or unexpected categories in real-world scenarios.

\begin{figure}[t]
\centering
\includegraphics[width=0.9\linewidth]{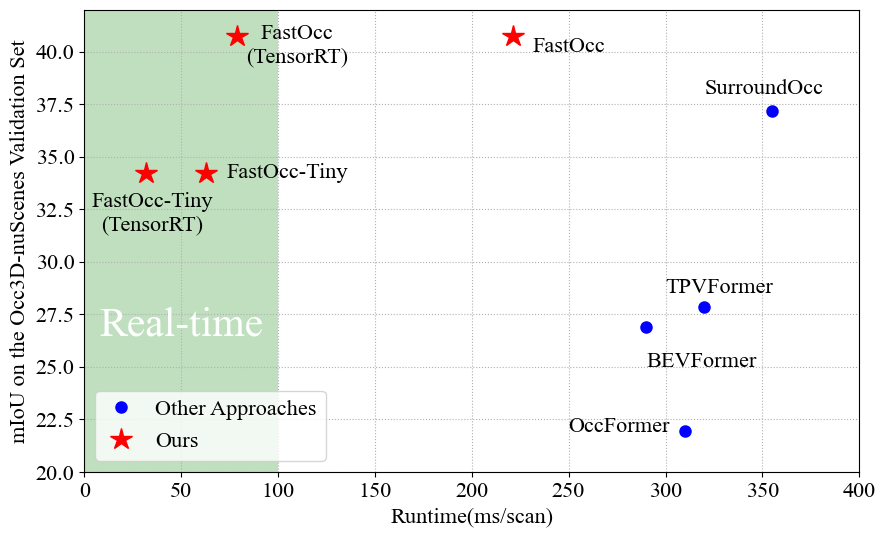}
\caption{Comparisons of the mIoU and runtime of various 3D occupancy prediction methods on the Occ3D-nuScenes~\cite{tianOcc3DLargeScale3D2023} validation set.}
\label{latency}
\end{figure}

Camera-based 3D occupancy prediction task takes the multi-camera images as inputs and estimates the occupancy status and semantic label of each 3D voxel of the entire surrounding. Unlike 3D object detection and other perception tasks, it provides denser perception results and demonstrates greater robustness against the weird objects~\cite{tianOcc3DLargeScale3D2023}, such as buses with bending connections or construction vehicles with long mechanical arms. Moreover, the voxel-based representation has the potential to be extended to various tasks, such as 3D semantic segmentation. At the same time, predicting the occupancy voxels is more efficient than reconstructing the whole 3D scene in detail because most autonomous driving tasks do not need over-elaborate details, such as tree leaves, windows of buildings, the texture of sidewalk tiles, and so on.

Despite the advantages mentioned above, 3D occupancy prediction is a highly challenging task that demands robustness, accuracy, and practical real-time efficiency. The previous works~\cite{huangTriPerspectiveViewVisionBased2023, tianOcc3DLargeScale3D2023, wangOpenOccupancyLargeScale2023, wangPanoOccUnifiedOccupancy2023, weiSurroundOccMultiCamera3D2023} investigated various aspects of 3D occupancy prediction tasks, including the feature representation, transformation from the image view to the voxel view, elaborate networks and ground-truth label generation, to improve the prediction accuracy. However, as shown in Fig.~\ref{latency}, many existing methods suffer from a significant computational burden during the prediction process, making them unsuitable for the real-time perception requirements, which is vital for autonomous driving.

To this end, we propose FastOcc, a new 3D occupancy prediction method with the real-time inference speed and competitive accuracy compared with the state-of-the-art approaches. The network effect and latency of different approaches are extensively evaluated and illustrated in the ablation study according to four parts, including the input image resolution, image backbone, view transformation, and occupancy prediction head. From these experimental results, it is observed that the 3D convolution or deconvolution used in the occupancy prediction head has considerable potential for optimizing the speed-accuracy trade-off. While most existing methods lift image features to 3D voxel features and straightly decode them in 3D representation, our proposed method first employs a fast approach to obtain volume features. Then the 3D form feature is collapsed to the 2D BEV representation and decoded in the BEV form. To address the absence of $z$-axis information in the BEV representation, a fast and simple interpolation sampling method is applied to extract 3D features with height information from the image features. Subsequently, the BEV features and the interpolated features are integrated for the final prediction results. Essentially, our method simplifies the process of a 3D perception task as the feature is compressed to BEV representation and decoded in 2D form, and then interpolated 3D features are employed to refine and enhance the 2D features. Supervision is applied both on BEV features and final voxel features. Our proposed method achieves state-of-the-art results with high efficiency compared to other methods. Furthermore, to adapt our method to the real-time perception requirements of autonomous driving, the network structure and setups are optimized and accelerated while ensuring precision. TensorRT SDK~\cite{vanholder2016efficient} is also employed for further acceleration.

Our contributions can be summarized as follows:
\begin{itemize}
    \item A detailed comparison of the network effect and latency is conducted on four parts in the occupancy prediction task, including the input image resolution, image backbone, view transformation, and occupancy prediction head. Results are presented in the ablation study.
    \item A novel efficient approach named FastOcc is proposed, which accelerates the 3D occupancy prediction process by simplifying 3D convolution blocks to a 2D BEV convolution network and completing the BEV features with the interpolated voxel features.
    \item FastOcc achieves the state-of-the-art mIoU of 40.75 while running much faster compared to other methods on the Occ3D-nuScenes~\cite{tianOcc3DLargeScale3D2023} dataset. The latency of a single inference is reduced to 63\,ms and can be further reduced to 32 \,ms with the TensorRT SDK~\cite{vanholder2016efficient} acceleration.
\end{itemize}

%% file: sections/related_work.tex
\section{RELATED WORK}



\subsection{Traditional Visual Perception}

In recent years, there has been a growing interest in the perception of autonomous vehicles to understand the surrounding environment. BEV perception\cite{harleySimpleBEVWhatReally2022, huangBEVDetHighperformanceMulticamera2021, liBEVDepthAcquisitionReliable2022, liBEVFormerLearningBird2022, philionLiftSplatShoot2020} has been one of the focal points. Various methods aimed to transform the individual feature representations from RGB cameras into a unified representation, which facilitates modeling of the surrounding environment. LSS\cite{philionLiftSplatShoot2020} estimated per-pixel depth and used the depth feature to place features at their estimated 3D locations. Simple-BEV\cite{harleySimpleBEVWhatReally2022} proposed to project the pre-defined 3D coordinates into images and rise bilinearly sampled features to 3D
volume grids. BEVFormer\cite{liBEVFormerLearningBird2022} used deformable attention operations to integrate image features into 3D grid coordinates.

3D object detection\cite{huangBEVDetHighperformanceMulticamera2021, wangDETR3D3DObject2021, luoDETR4DDirectMultiView2022, wangFCOS3DFullyConvolutional2021} has emerged as a simple and effective approach for perception, leveraging input from surround-view RGB cameras. Various works \cite{youPseudoLiDARAccurateDepth2020, wangFCOS3DFullyConvolutional2021, liuPETRv2UnifiedFramework2022, wangDETR3D3DObject2021} have reached great effect on this task, which allows for the accurate estimation of objects using 3D bounding box with dimensions, positions, and orientation. The bounding box has been widely accepted as a suitable representation for autonomous driving tasks, especially for objects in traffic environments that exhibit rigid body attributes, such as vehicles. However, some objects with unique shapes and irregular structures are not well-suited for this format.

\subsection{3D Occupancy Prediction}
3D occupancy perception\cite{caoMonoSceneMonocular3D2022, huangTriPerspectiveViewVisionBased2023, liFBOCC3DOccupancy2023, tianOcc3DLargeScale3D2023, weiSurroundOccMultiCamera3D2023} is a task that can obtain more detailed scene perception results while demonstrating good scalability and adaptability to downstream tasks. The pioneering Monoscene\cite{caoMonoSceneMonocular3D2022} utilized a monocular camera as input for semantic scene completion. It employed a continuous 2D-3D UNet\cite{ronnebergerUNetConvolutionalNetworks2015} to map the image feature to a 3D representation. However, due to the monocular perspective limitation, inferring fine-grained and accurate results with a simple framework is challenging and vulnerable to occlusion, distortion, and ghosting issues. TPVFormer\cite{huangTriPerspectiveViewVisionBased2023} incorporated surround multi-camera input and lifted features to a tri-perspective view space using a transformer-based approach. As it relied on sparse LiDAR points for supervision, the predicted results were also sparse. SurroundOcc\cite{weiSurroundOccMultiCamera3D2023} generated 3D voxel features at multiple scales using a transformer-based approach and combined them through deconvolutional upsampling. It also proposed a pipeline to obtain dense semantic occupancy supervision from sparse LiDAR information, resulting in a dense prediction. CTF-Occ\cite{tianOcc3DLargeScale3D2023} gradually refined the 3D voxel features from various scales in a coarse-to-fine manner and constructed a dense visibility-aware benchmark. However, the prediction process in these methods is time-consuming and far from the real-time perception requirements of autonomous driving. For example, the network of SurroundOcc\cite{weiSurroundOccMultiCamera3D2023} takes more than 300\,ms for a single inference. While most methods directly enhance the feature transformation from images to dense 3D voxel representations using carefully designed approaches to achieve better results, our approach converts image features to BEV features in a straightforward manner and employs a fast interpolation method to complement the missing height dimension features of BEV, resulting in equally accurate occupancy prediction results with significantly reduced computational overhead.

%% file: sections/occupancy_prediction.tex
\section{Methodology}

\begin{figure*}[ht]
\centering
\includegraphics[width=0.9\linewidth]{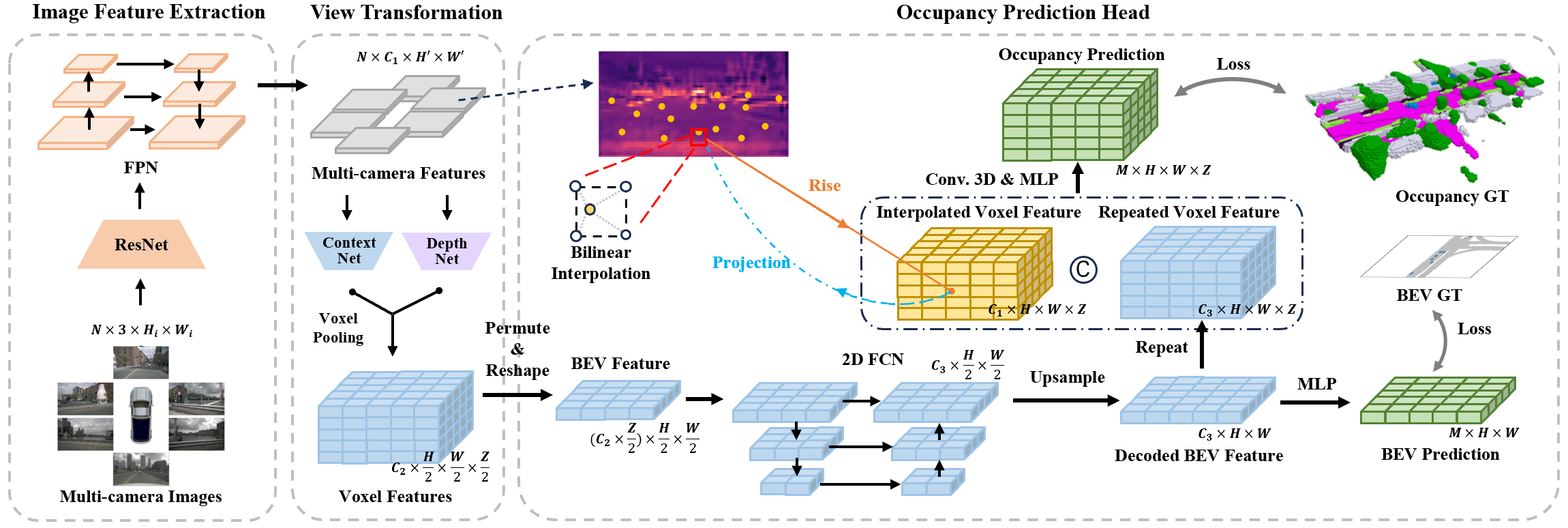}
\caption{The pipeline of the proposed method. First, multi-camera features are extracted from image inputs with a backbone network. Then image features are transformed to the 3D space following the LSS\cite{philionLiftSplatShoot2020} strategy. The voxel feature is collapsed to the BEV form and decoded in the 2D representation. Subsequently, the BEV features are upsampled, repeated, and supplemented with the voxel features interpolated from image features. BEV semantic segmentation is supervised as an auxiliary loss.}
\label{method_pipeine}
\end{figure*}

In this section, first, we illustrate the visual 3D occupancy prediction task and provide a formulaic expression of the entire process in \ref{Problem Formulation}. Subsequently, as shown in Fig. \ref{method_pipeine},  the pipeline of the proposed FastOcc can be divided into three parts, including image feature extraction, view transformation, and occupancy prediction head. \ref{Image Feature Extraction} shows the employed feature extraction backbone. In \ref{View Transformation}, widely-used 2D-to-3D view transformation methods are evaluated and the strategy used in our approach is illustrated. Most importantly, our novel occupancy prediction head is illustrated in \ref{Occupancy Prediction Head}, where the 3D convolution blocks are simplified by a 2D BEV convolution network, and 2D features are fused with the interpolated voxel features for further fine-tuning. \ref{Loss Function} introduces the training loss function.

\subsection{Problem Formulation}
\label{Problem Formulation}

In this work, the 3D surrounding scene to be predicted is divided by voxels. Assuming that the autonomous ego is placed at the origin of the real-world coordinates, the scene perception range is denoted as $[H_{s}, W_{s}, Z_{s}, H_{e}, W_{e}, Z_{e}]$. Given that the shape of 3D volume grids is $[ H, W, Z]$, each voxel $v$ has the shape of
\begin{equation}
[\frac{W_{e}-W_{s}}{W}, \frac{H_{e}-H_{s}}{H}, \frac{Z_{e}-Z_{s}}{Z}],
\end{equation}
and the semantic occupancy labels can be defined as $\mathbf{Y}^* \in \mathbb{R} ^{M\times H\times W\times Z}$, where $M$ is the number of semantic labels, including the unoccupied voxels denoted as \textit{empty}. Taking multi-camera images $\mathbf{X} = \{\mathbf{X}^1, \mathbf{X}^2, \cdots, \mathbf{X}^N\}$ from $N$ cameras as input, a neural network $\mathcal{G}$ is developed to tackle the semantic occupancy prediction task, which is represented as:

\begin{equation}
\mathbf{Y} = \mathcal{G}(\mathbf{X}^1, \mathbf{X}^2, \cdots, \mathbf{X}^N),
\end{equation}
where $\mathbf{Y}$ is the predicted result.

\subsection{Image Feature Extraction}
\label{Image Feature Extraction}

The image feature extraction process takes multi-camera images $\mathbf{X} \in \mathbb{R}^{N \times 3\times H_i \times W_i}$ as inputs, where $[H_i, W_i]$ is the shape of input images. Then a UNet-like\cite{ronnebergerUNetConvolutionalNetworks2015} backbone is employed to extract multi-camera features $\mathbf{F} = \{\mathbf{F}^1, \mathbf{F}^2, \cdots, \mathbf{F}^N\}$. In our implementation, ResNet-like\cite{heDeepResidualLearning2016} blocks are employed to encode image features to $1/16$ of the origin shape and the feature pyramid network (FPN)\cite{linFeaturePyramidNetworks2017} is applied to aggregate features into scale $[H', W']$. The output feature can be denoted as $\mathbf{F} \in \mathbb{R}^{N \times C_1\times H' \times W'}$.

\subsection{View Transformation}
\label{View Transformation}

In the view transformation process, image features $\mathbf{F}$ from multiple cameras are lifted to a unified 3D form to represent the 3D scene uniquely. The transformed feature can be denoted as $\mathbf{V_B} \in \mathbb{R}^{C_2 \times \frac{H}{2} \times \frac{W}{2} \times \frac{Z}{2}}$, where $C_2$ is the embedding dim, and to lower the cost, features are transformed to a rather coarse grid size $[\frac{H}{2}, \frac{W}{2}, \frac{Z}{2}]$. Many previous occupancy prediction methods\cite{liVoxFormerSparseVoxel2023, weiSurroundOccMultiCamera3D2023, tianOcc3DLargeScale3D2023} build 3D volume queries and apply the cross-view attention\cite{liBEVFormerLearningBird2022} to integrate the multi-view 2D image features into 3D space. However, for high efficiency, the principle proposed by LSS\cite{philionLiftSplatShoot2020} is employed as our view transformation strategy. The LSS\cite{philionLiftSplatShoot2020} approach estimates the depth and context features simultaneously and applies a voxel-pooling mechanic to integrate the 2D features into 3D representation.
Moreover, we adopt the BEVDepth\cite{liBEVDepthAcquisitionReliable2022}, which introduces point clouds to supervise the depth feature predicted by the depth net of LSS\cite{philionLiftSplatShoot2020}. By estimating the depth of each pixel, the image features are projected with depth uncertainty accounted for. The transformation strategy, which applies the depth supervision together with the depth-context correspondence, is demonstrated to have a better performance and faster speed in our experiments.

\subsection{Occupancy Prediction Head}
\label{Occupancy Prediction Head}

To get the 3D prediction output efficiently and effectively, the original 3D feature decoding process is replaced by a residual-like architecture, which is composed of the BEV feature decoding process, the image feature interpolation sampling for compensating the $z$-axis information, and the final feature integration. These components are introduced as follows.

\begin{figure}[t]
\centering
\subfloat[2D FCN]{
		\includegraphics[width=0.55\linewidth]{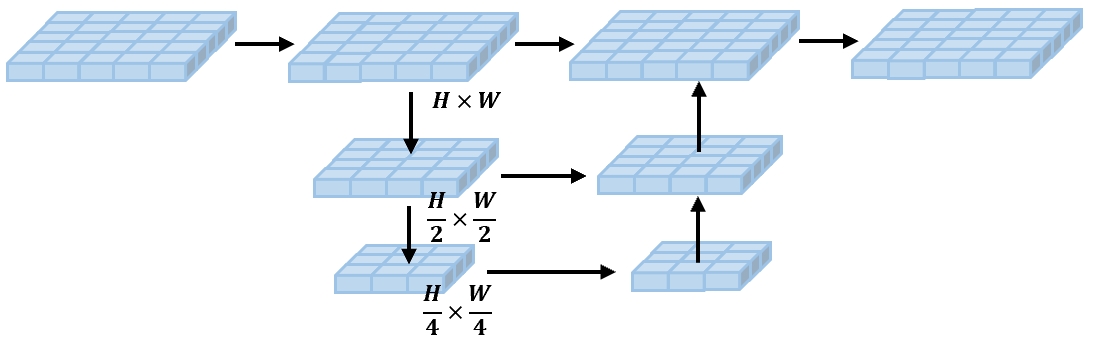}}\\
\subfloat[3D FCN]{
		\includegraphics[width=0.6\linewidth]{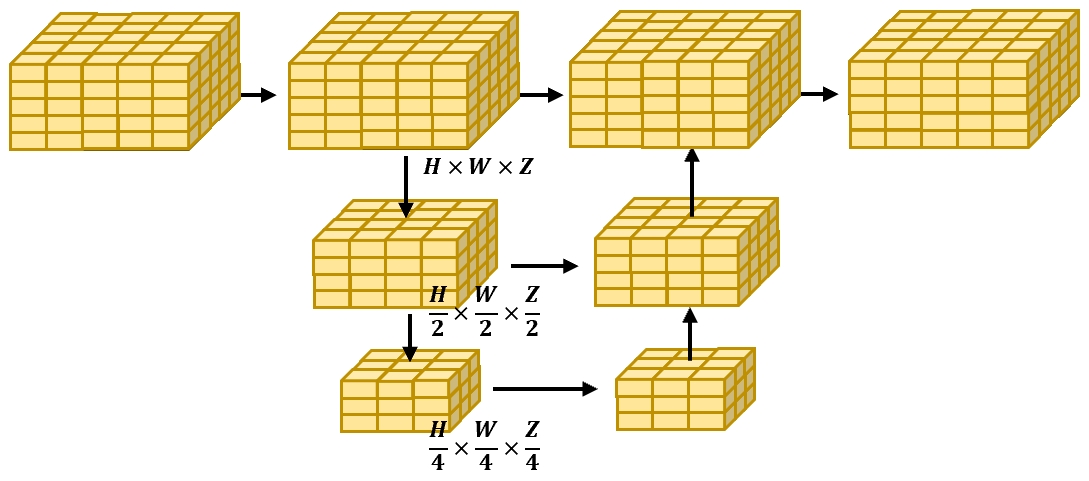}}
\caption{The comparison of applying 2D FCN and 3D FCN. It is obvious that 2D FCN is highly efficient in terms of time and memory cost.}
\label{fcn_comparison}
\end{figure}

\textbf{BEV Feature Decoding.} 
Most of the existing methods directly decode the volume features in 3D form. Taking the 3D fully convolutional network (FCN) as an example, for the $j^{\text{th}}$ 3D convolution layer, the number of floating point operations (FLOPs) can be calculated as 
\begin{equation}
FLOPs_j^{3D} = C_j^{in} \times {k_j}^3 \times C_j^{out} \times H_j \times W_j \times Z_j,
\end{equation}
where in layer $j$, $C_j^{in}$ is the number of the input channels, $k_j$ is the convolution kernel size, $C_j^{out}$ is the number of the output channels, and $[H_j, W_j, Z_j]$ is the shape of the 3D feature map.

Compared with straightly decoding the lifted voxel feature in 3D space, the proposed method employs a lightweight 2D BEV decoder. Given the previous view transformation outputs $\mathbf{V_B} \in \mathbb{R}^{c_2\times\frac{H}{2}\times\frac{W}{2}\times\frac{Z}{2}}$, the proposed method first combine the $z$ dim of 3D voxel features $\mathbf{V_B}$ with its embedding channel to get the 2D BEV features $\mathbf{B'} \in \mathbb{R}^{(C_2\times \frac{Z}{2}) \times \frac{H}{2} \times \frac{W}{2}}$. Then $\mathbf{B'}$ is decoded with a 2D FCN to the BEV feature $\mathbf{B} \in \mathbb{R}^{C_3 \times \frac{H}{2} \times \frac{W}{2}}$, as shown in Fig. \ref{fcn_comparison}. This reduces the computational complexity to a large extent. The FLOPs in each 2D convolution layer $j$ can be calculated as
\begin{equation}
FLOPs_j^{2D} = C_j^{in} \times {k_j}^2 \times C_j^{out} \times H_j \times W_j.
\end{equation}
Consequently, in the first layer, $C_1^{in} = C_2 \times \frac{Z}{2}$, the 2D convolution layer is theoretically $k$ times faster than the 3D convolution layer. In the subsequent layer $j\,(j>1)$, the 2D convolution layers is $s_j$ times faster than 3D ones, $s_j$ can be computed as
\begin{equation}
s_j = \frac{FLOPs_j^{3D}}{FLOPs_j^{2D}} = k_j \times Z_j.
\end{equation}

\begin{figure}[t]
\centering
\includegraphics[width=0.85\linewidth]{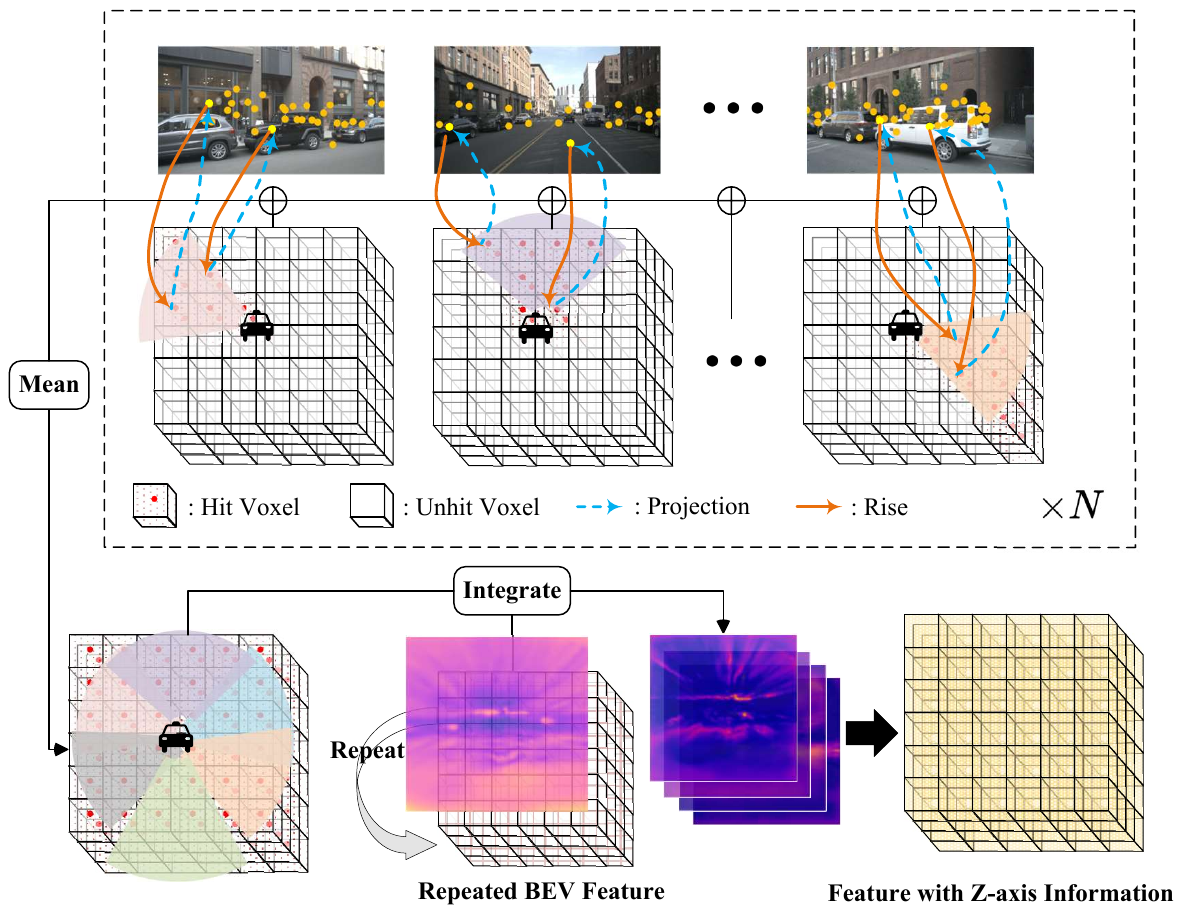}
\caption{In the upper dashed box, the volume grids are projected to multiple perspective images. Features of hit voxels on the sub-pixels are bilinearly interpolated and lifted to corresponding 3D space. Below, the absence of the $z$-axis of repeated BEV features can be completed by the interpolated features.}
\label{interpolation}
\end{figure}

\textbf{Image Feature Interpolation Sampling.} To augment the absent $z$-axis information in the BEV form and minimize the computational complexity, a simple and efficient approach is designed to acquire 3D features.

To be more specific, first, a 3D volume coordinate is created according to the voxel space shape $[H, W, Z]$ and assigned to the ego coordinate, defined as $C_{ego}$. Then the transformation from the ego to the image can be computed as $T_{e2i} = T_{c2i} \times T_{e2c}$, where $T_{c2i}$ is the camera intrinsic matrix and $T_{e2c}$ is the transformation from ego car to camera. The coordinate $C_{ego}$ is projected to the images to get the correspondence between the grid coordinate and perspective-view features, and the projected grid can be defined as $C_{image} = T_{e2i} \times C_{ego}$. After that, points that exceed the image range or have a negative depth are filtered out. Subsequently, we apply bilinear sampling to interpolate features from projected sub-pixel coordinates on multiple cameras and compute the mean value after masking out unobserved voxels. Fig. \ref{interpolation} illustrates the detailed process. The FLOPs of the interpolation sampling process is
\begin{equation}
FLOPs^{inter}=4N\times C\times H\times W\times Z,
\end{equation}
where 4 neighbor pixels are referred for bilinear sampling sub-pixel features with dim $C$ from $N$ cameras.

\begin{table*}[ht]
\centering
\renewcommand{\arraystretch}{1}
\setlength{\tabcolsep}{2.5pt}
\caption{3D semantic occupancy prediction performance on the validation set of Occ3D-nuScenes\cite{tianOcc3DLargeScale3D2023}. For a fair comparison, we train SurroundOcc\cite{weiSurroundOccMultiCamera3D2023} on the Occ3D-nuScenes dataset with its origin setups, denoted as SurroundOcc*.}
\begin{tabular}{l|ccccccccccccccccc|c}
\hline
Method  & \rotatebox{90}{\textcolor{black}{$\blacksquare$} others} & \rotatebox{90}{\textcolor[RGB]{255,120,50}{$\blacksquare$} barrier} & \rotatebox{90}{\textcolor[RGB]{255,192,203}{$\blacksquare$} bicycle} & \rotatebox{90}{\textcolor[RGB]{255,255,0}{$\blacksquare$} bus} & \rotatebox{90}{\textcolor[RGB]{0,150,245}{$\blacksquare$} car} & \rotatebox{90}{\textcolor[RGB]{0,255,255}{$\blacksquare$} const. veh.} & \rotatebox{90}{\textcolor[RGB]{255,127,0}{$\blacksquare$} motorcycle} & \rotatebox{90}{\textcolor[RGB]{255,0,0}{$\blacksquare$} pedestrian} & \rotatebox{90}{\textcolor[RGB]{255,240,150}{$\blacksquare$} traffic cone } & \rotatebox{90}{\textcolor[RGB]{135,60,0}{$\blacksquare$} trailer} & \rotatebox{90}{\textcolor[RGB]{160,32,240}{$\blacksquare$} truck} & \rotatebox{90}{\textcolor[RGB]{255,0,255}{$\blacksquare$} drive. surf.} & \rotatebox{90}{\textcolor[RGB]{139,137,137}{$\blacksquare$} other flat} & \rotatebox{90}{\textcolor[RGB]{75,0,75}{$\blacksquare$} sidewalk} & \rotatebox{90}{\textcolor[RGB]{150,240,80}{$\blacksquare$} terrain} & \rotatebox{90}{\textcolor[RGB]{230,230,250}{$\blacksquare$} manmade} & \rotatebox{90}{\textcolor[RGB]{0,175,0}{$\blacksquare$} vegetation} & \rotatebox{90}{mIoU} \\ \hline
MonoScene\cite{caoMonoSceneMonocular3D2022} &1.75&7.23&4.26&4.93&9.38&5.67&3.98&3.01&5.90&4.45&7.17&14.91&6.32&7.92&7.43&1.01&7.65&6.06\\
TPVFormer\cite{huangTriPerspectiveViewVisionBased2023}  &7.22&38.90&13.67&40.78&45.90&17.23&19.99&18.85&14.30&26.69&34.17&55.65&35.47&37.55&30.70&19.40&16.78&27.83 \\
BEVDet\cite{huangBEVDetHighperformanceMulticamera2021}  &4.39&30.31&0.23&32.36&34.47&12.97&10.34&10.36&6.26&8.93&23.65&52.27&24.61&26.06&22.31&15.04&15.10&19.38 \\
OccFormer\cite{zhangOccFormerDualpathTransformer2023}  &5.94&30.29&12.32&34.40&39.17&14.44&16.45&17.22&9.27&13.90&26.36&50.99&30.96&34.66&22.73&6.76&6.97&21.93 \\
BEVFormer\cite{liBEVFormerLearningBird2022}  &5.85&37.83&17.87&40.44&42.43&7.36&23.88&21.81&20.98&22.38&30.70&55.35&28.36&36.0&28.06&20.04&17.69&26.88 \\
CTF-Occ\cite{tianOcc3DLargeScale3D2023}  &8.09&39.33&20.56&38.29&42.24&16.93&24.52&22.72&21.05&22.98&31.11&53.33&33.84&37.98&33.23&20.79&18.0&28.53 \\
SurroundOcc*\cite{weiSurroundOccMultiCamera3D2023} &8.97&\textbf{46.33}&17.08&\textbf{46.54}&52.01&20.05&21.47&23.52&18.67&\textbf{31.51}&37.56&81.91&41.64&50.76&\textbf{53.93}&\textbf{42.91}&\textbf{37.16}&37.18 \\
FastOcc(Ours) &\textbf{12.06}&43.53&\textbf{28.04}&44.80&\textbf{52.16}&\textbf{22.96}&\textbf{29.14}&\textbf{29.68}&\textbf{26.98}&30.81&\textbf{38.44}&\textbf{82.04}&\textbf{41.93}&\textbf{51.92}&53.71&41.04&35.49&\textbf{39.21}      \\ \hline
SurroundOcc*-TTA\cite{weiSurroundOccMultiCamera3D2023} &9.42&43.61&19.57&\textbf{47.66}&53.77&21.26&22.35&24.48&19.36&32.96&39.06&83.15&43.26&52.35&55.35&\textbf{43.27}&\textbf{38.02}&38.69 \\
FastOcc-TTA(Ours) &\textbf{12.86}&\textbf{46.58}&\textbf{29.93}&46.07&\textbf{54.09}&\textbf{23.74}&\textbf{31.10}&\textbf{30.68}&\textbf{28.52}&\textbf{33.08}&\textbf{39.69}&\textbf{83.33}&\textbf{44.65}&\textbf{53.90}&\textbf{55.46}&42.61&36.50&\textbf{40.75}      \\ \hline
\end{tabular}
\label{sota_comparison}
\end{table*}

\textbf{Feature Integration.} To integrate the 2D BEV feature with interpolated 3D voxel feature, decoded BEV features $\mathbf{B}$ at scale $[\frac{H}{2}, \frac{W}{2}]$ are upsampled to a fine-grained scale $[H, W]$ and repeated at the $z$-axis, denoted as $\mathbf{B_z} \in \mathbb{R}^{C_3 \times H \times W\times Z}$. The interpolated voxel feature $\mathbf{P} \in \mathbb{R}^{C_1 \times H \times W\times Z}$ is obtained in a fast manner directly at the fine-grained scale with more detailed information. $\mathbf{B_z}$ and $\mathbf{P}$ are concatenated together and integrated by a convolution layer to get the output voxel feature $\mathbf{V}$.

Moreover, to ensure that the decoded BEV feature $\mathbf{B}$ contains enough information for further fine-tuning, it is processed by a UNet-like\cite{ronnebergerUNetConvolutionalNetworks2015} semantic segmentation head and supervised by the BEV ground truth $\mathbf{B_{gt}} \in \mathbb{R}^{M \times H \times W}$. To generate the BEV ground truth $\mathbf{B_{gt}} \in \mathbb{R}^{M \times H \times W}$ from occupancy ground truth $\mathbf{V_{gt}} \in \mathbb{R} ^{M\times H\times W\times Z}$, we simply count the voxels occupied by each class at the $z$-axis and assign the BEV grid as occupied by each class using a binary multi-class vector.



Rather than simply repeating BEV features to 3D form, which results in redundancy on the $z$-axis, integrating with the interpolated voxel features incorporates multiple perspective images and achieves better scene understanding, as shown in Fig. \ref{interpolation}.

For the entire occupancy prediction head. If a 3D FCN is applied, the computation complexity is of $O(k^3C_{in}C_{out}HWZ)$. In our method, the cost of interpolation sampling is much less than multiple convolution layers, consequently, the computational complexity is dominated by $O(k^2C_{in}C_{out}HW)$.

\subsection{Loss Function}
\label{Loss Function}
To train the model, we apply the focal loss $L_f$ following M2BEV\cite{xieBEVMultiCameraJoint2022}, the affinity loss $L_{sem}$, $L_{geo}$, and dice loss $L_{dice}$ introduced in MonoScene\cite{caoMonoSceneMonocular3D2022}, the lovasz-softmax loss $L_{ls}$ from OpenOccupancy\cite{wangOpenOccupancyLargeScale2023}. As mentioned above, to ensure that the features are transformed of high quality, we supervise the perspective depth with $L_{d}$ and BEV feature map with binary cross-entropy loss $L_{b}$. The final loss is composed of:

\begin{equation}
Loss= L_{f} + L_{sem} + L_{geo} + L_{dice} + L_{ls} + L_{d} + L_{b}.
\end{equation}

%% file: sections/experiment.tex
\section{EXPERIMENTS}

\begin{table}[t]
\centering
\renewcommand{\arraystretch}{1.2}
\setlength{\tabcolsep}{2.8pt}
\caption{Comparisons of the mIoU and latency of the proposed components. The SurroundOcc$^*$ is progressively refined to the proposed FastOcc. The input image size is $640\times 1600$ and image backbone is the ResNet-101.}
\begin{tabular}{l|cc|cc}
\hline
Method  & \makecell{View\\ Transformation} & \makecell{Occupancy \\Prediction Head} & mIoU & Latency(ms) \\ 
\hline
SurroundOcc$^*$ & BEVFormer\cite{liBEVFormerLearningBird2022} & Deconv. & 37.18 & 355 \\
Baseline & LSS\cite{philionLiftSplatShoot2020} & Deconv. & 38.44 & \underline{306} \\
Baseline$^+$ & LSS\cite{philionLiftSplatShoot2020} & 3D FCN & \textbf{41.02} & 342 \\
FastOcc & LSS\cite{philionLiftSplatShoot2020} & 2D FCN & \underline{40.75} & \textbf{221} \\
\hline
\end{tabular}
\label{method_comparison}
\end{table}


\begin{table}[t]
\centering
\renewcommand{\arraystretch}{1.2}
\caption{The ablation study of the image backbones and input resolutions. TRT means the acceleration of the TensorRT SDK\cite{vanholder2016efficient}.}
\setlength{\tabcolsep}{1.5pt}
\scriptsize
\begin{tabular}{ll|cc|cc}
\hline
\multicolumn{2}{c|}{\footnotesize Method}& {\footnotesize Backbone}&{\footnotesize Input Res.}& {\footnotesize mIoU} & \makecell{\footnotesize Latency(ms)\\ \hline 2D\ /\ 2D-to-3D\ /\ 3D\ /\ Total} \\ \hline
\multicolumn{1}{l|}{\multirow{3}{*}{\rotatebox{90}{\scriptsize Pytorch}}} & {\scriptsize FastOcc-Tiny} & ResNet-50 & $320\times 800$ & 34.21 & 26.32 / 3.59 / 32.89 / \textbf{62.80}\\
\multicolumn{1}{l|}{} & {\scriptsize FastOcc-Small} & ResNet-101 & $320\times 800$ & 37.21 & 53.86 / 3.59 / 32.89 / 90.34 \\
\multicolumn{1}{l|}{} & {\scriptsize FastOcc} & ResNet-101 & $640\times 1600$ & \textbf{40.75} & 176.82 / 11.57 / 32.89 / 221.28\\ \hline
\multicolumn{1}{c|}{\multirow{3}{*}[3pt]{\rotatebox{90}{\scriptsize TRT}}} & {\scriptsize FastOcc-Tiny} & ResNet-50 & $320\times 800$ & 34.21 & 14.11 / 1.80 / 16.02 / \textbf{31.94}\\
\multicolumn{1}{l|}{} & {\scriptsize FastOcc} & ResNet-101 & $640\times 1600$ & \textbf{40.75} & 58.42 / 3.62 / 16.21 / 78.25\\ \hline
\end{tabular}

\label{acceleration}
\end{table}

\begin{table}[t]
\centering
\renewcommand{\arraystretch}{1.0}
\caption{The ablation study of the BEV supervision and interpolated feature fusion. The decoded BEV features are straightly repeated and regressed to occupancy if interpolated features are not fused with.}
\begin{tabular}{cc|c}
\hline
BEV Supervision & \makecell{Interpolated \\ Feature Fusion} & mIoU \\ \hline
$\surd$ & - & 31.67  \\
- & $\surd$ & 33.08 \\
$\surd$ & $\surd$ & \textbf{34.21}  \\ \hline
\end{tabular}
\label{ablation}
\end{table}

\begin{figure*}[ht]
\centering
\includegraphics[width=0.85\linewidth]{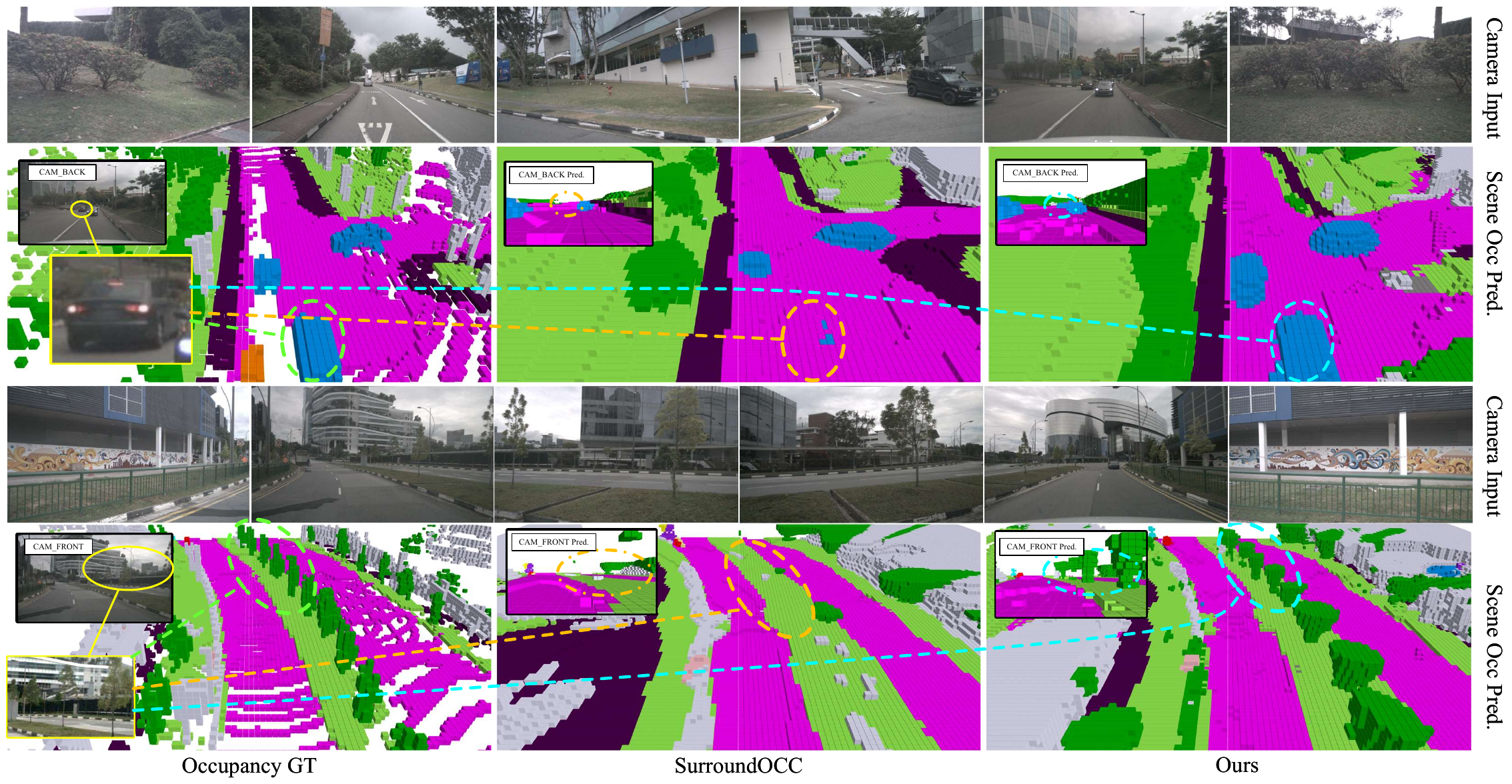}
\caption{Visualization of the occupancy prediction results on the validation set of Occ3D-nuScenes\cite{tianOcc3DLargeScale3D2023}.}
\label{occupancy_vis}
\end{figure*}

\subsection{Experimental Setups}
\textbf{Dataset and Evaluation Metrics.}
Occ3D-nuScenes\cite{caesarNuScenesMultimodalDataset2020, tianOcc3DLargeScale3D2023} provides the ground truth of a voxelized representation of the 3D space, with the occupancy state and semantic labels jointly estimated. The benchmark contains 28,130 train samples, 6,019 validation samples, and 6,008 test samples. The perception range is $[-40\,m, -40\,m, -1\,m, 40\,m, 40\,m, 5.4\,m]$ and is divided by voxels with size $0.4\,m$. The voxels are classified into 18 semantic categories. 

For evaluation, following the previous works\cite{weiSurroundOccMultiCamera3D2023, liFBOCC3DOccupancy2023, huangTriPerspectiveViewVisionBased2023, liVoxFormerSparseVoxel2023}, the mean intersection over union (mIoU) of all semantic classes is employed for the 3D semantic occupancy prediction task.

\textbf{Implementation Details.}
For our best result, ResNet-101\cite{heDeepResidualLearning2016} pretrained on FCOS3D\cite{wangFCOS3DFullyConvolutional2021} is employed as the image backbone, and the input image size is cropped to $640 \times 1600$. The employed FPN\cite{linFeaturePyramidNetworks2017} has three levels of layers. Image features are transformed to voxel features with shape $[100\times 100\times 8]$. The collapsed BEV feature has the shape of $[100\times 100]$, and the 2D FCN decoder is composed of a ResNet-18\cite{heDeepResidualLearning2016} and a 3-level FPN\cite{linFeaturePyramidNetworks2017}. The decoded BEV features are upsampled and repeated to $[200\times 200\times 16]$, which is the same shape as interpolated features. The AdamW\cite{loshchilovDecoupledWeightDecay2019} optimizer and cosine annealing\cite{loshchilovSGDRStochasticGradient2017} learning rate scheduler with a warm-up is employed, and the learning rate is initialized to $2e-4$. Data augmentation on both input images and 3D voxels is employed. Test-time augmentation and camera masks that ignore those invisible voxels are also applied. Temporal information from the previous 16 frames is considered for better results. The experiments are conducted on four Tesla V100 GPUs.

\subsection{Evaluation Comparisons}

Table \ref{sota_comparison} illustrates the comparison of mIoU scores among our method and other relevant approaches for the 3D occupancy prediction task. It is evident that our method achieves high performance on mIoU and most of the categories. Fig. \ref{occupancy_vis} shows the prediction results of FastOcc compared with SurroundOcc\cite{weiSurroundOccMultiCamera3D2023}. It is obvious that FastOcc fills the blank grids of the ground truth in a more reasonable manner and avoids perception failures on distant cars and blurry trees.

\subsection{Ablation Study}

\textbf{Effects of View Transformation and Occupancy Prediction Head.} The transformation method to lift 2D features to 3D space has always been a popular topic. We compare the efficiency and the resulting mIoU scores of the transformer-based method\cite{liBEVFormerLearningBird2022} and LSS\cite{philionLiftSplatShoot2020} strategy on our baseline work. Table \ref{method_comparison} shows the results. SurroundOcc$^*$\cite{weiSurroundOccMultiCamera3D2023} employs multi-scale cross-view attention\cite{liBEVFormerLearningBird2022} as the view transformation method and decodes the features using 3D deconvolution network. We implement the Baseline applying the LSS\cite{philionLiftSplatShoot2020} strategy following \cite{liFBBEVBEVRepresentation2023}. Compared with SurroundOcc\cite{weiSurroundOccMultiCamera3D2023}, the Baseline model results in better results with faster speed. Moreover, the occupancy prediction head is ablated to show the efficiency of the proposed method. In Baseline$^+$, a 3D FCN is applied to get better results compared to the multi-scale deconvolution network used in Baseline, but the computation cost increases obviously. In FastOcc, the 2D FCN network is used as the occupancy prediction head, which retains the mIoU with a much faster inference speed. From the comparisons, it is obvious that the depth-supervised LSS\cite{philionLiftSplatShoot2020} and 2D FCN with interpolated features completion present both effectiveness and efficiency.

\textbf{Effects of Input Resolution and Image Backbone.}
We also evaluate the impact of the input image resolutions and image backbones. As shown in Tabel~\ref{acceleration}, both higher image resolution and stronger image backbone lead to more accurate results (higher mIoU). Besides, the proposed FastOcc is further accelerated by the TensorRT SDK\cite{vanholder2016efficient}. Specifically, FastOcc-Tiny and FastOcc run 31.94\,ms and 78.25\,ms, respectively, to meet the real-time inference requirement.

\textbf{Effects of BEV Supervision and Interpolation.}
Recovering the complete 3D voxel information from the 2D BEV features is a challenging task since the $z$-axis is absent. To tackle this problem, we propose two strategies: 1) the BEV supervision imposes the 3D information on the 2D BEV features; 2) the interpolated voxel features sampling from the images serve as a supplement. The results in Table~\ref{ablation} demonstrate the effectiveness of the two strategies.

%% file: sections/conclusions.tex
\section{CONCLUSIONS}

In this paper, FastOcc is proposed for efficient 3D semantic occupancy prediction. 3D voxel features are compressed to be 2D BEV features after view transformation, where a 2D FCN is applied for efficient feature extraction. Subsequently, the absent $z$-axis of the BEV features is compensated by the interpolated voxel features from the image, resulting in the complete 3D voxel information with efficiency. Comparisons with other methods on the Occ3D-nuScenes\cite{tianOcc3DLargeScale3D2023} dataset demonstrate the advantages of the proposed components. The proposed FastOcc achieves a leading mIoU of 40.75 and the FastOcc-Tiny runs 32\,ms with the TensorRT SDK\cite{vanholder2016efficient} acceleration.